# AcoustoBots: A Swarm of Robots for Acoustophoretic Multimodal Interactions

**Narsimlu Kemsaram** [1,*], **James Hardwick** [1], **Jincheng Wang** [1], **Bonot Gautam** [1], **Ceylan Besevli** [1], **Giorgos Christopoulos** [1], **Sourabh Dogra** [1], **Lei Gao** [1], **Akin Delibasi** [1], **Diego Martinez Plasencia** [1], **Orestis Georgiou** [2], **Marianna Obrist** [1], **Ryuji Hirayama** [1], and **Sriram Subramanian** [1]

[1]*Department of Computer Science, University College London, United Kingdom*
[2]*Ultraleap Limited, Bristol, United Kingdom*

Correspondence*:
Narsimlu Kemsaram
n.kemsaram@ucl.ac.uk

## ABSTRACT

Acoustophoresis has enabled novel interaction capabilities, such as levitation, volumetric displays, mid-air haptic feedback, and directional sound generation, to open new forms of multimodal interactions. However, its traditional implementation as a singular static unit limits its dynamic range and application versatility. This paper introduces "AcoustoBots" - a novel convergence of acoustophoresis with a movable and reconfigurable phased array of transducers for enhanced application versatility. We mount a phased array of transducers on a swarm of robots to harness the benefits of multiple mobile acoustophoretic units. This offers a more flexible and interactive platform that enables a swarm of acoustophoretic multimodal interactions. Our novel AcoustoBots design includes a hinge actuation system that controls the orientation of the mounted phased array of transducers to achieve high flexibility in a swarm of acoustophoretic multimodal interactions. In addition, we designed a BeadDispenserBot that can deliver particles to trapping locations, which automates the acoustic levitation interaction. These attributes allow AcoustoBots to independently work for a common cause and interchange between modalities, allowing for novel augmentations (e.g., a swarm of haptics, audio, and levitation) and bilateral interactions with users in an expanded interaction area. We detail our design considerations, challenges, and methodological approach to extend acoustophoretic central control in distributed settings. This work demonstrates a scalable acoustic control framework with two mobile robots, laying the groundwork for future deployment in larger robotic swarms. Finally, we characterize the performance of our AcoustoBots and explore the potential interactive scenarios they can enable.

## 1 INTRODUCTION

Acoustophoresis, the use of sound waves to manipulate and move objects in mid-air, has enabled a diverse set of contactless autonomous systems that are transforming how people control and interact with tangible materials in the real world. For instance, orchestrating ultrasonic waves to levitate multiple small particles, has been explored for food levitation (Vi et al., 2017), 3D printing (Ezcurdia et al., 2022), and data physicalisation (Gao et al., 2023). Meanwhile, high-speed manipulation of levitated particles has also created volumetric displays via the persistence of vision (PoV) effect (Hirayama et al., 2019), (Plasencia et al., 2020). The same acoustophoretic principles have been extended to haptic feedback in mid-air, providing tactile experiences to users, and enabling interaction with virtual objects in free space (Carter et al., 2013). Furthermore, directional generation of audible sound uses the localized precision of acoustophoresis to project audio to specific targets (Ochiai et al., 2017), moving toward eliminating personal auditory devices. Taken together, these capabilities underscore the pivotal role of acoustophoresis in advancing the paradigms of autonomous robotic platforms, synthesizing multimodalities in one platform by one technical principle.

Although acoustophoresis is introducing novel opportunities for contactless robotic manipulation and multimodal interaction, its traditional implementation is limited by a single static device that lacks scalability, mobility, flexibility, modularity, and versatility in various environments (Marshall et al., 2012), (Hirayama et al., 2019), (Fushimi et al., 2019), (Plasencia et al., 2020), (Gao et al., 2023). For example, commercial ultrasound haptics are currently confined in a fixed, small interaction area (e.g., 63 × 48 × 48 cm for Ultraleap devices) (Faridan et al., 2022), restricting dynamic interactions in expansive environments. Similarly, singular directional audio setups often fail to adaptively target multiple listeners in a tabletop system, often compromising auditory precision and immersion. These limitations hinder the integration of acoustophoretic platforms in realistic application environments. When it comes to levitation and volumetric displays, static configurations also hinder the dynamic projection of 3D content across various tangible locations on a tabletop or support the handover of levitated content from one board to the other, ultimately reducing the richness of interactive experiences. Meanwhile, the acoustophoretic system consists of complex algorithms to shape the sound field and precise control of hardware components, so it is not possible to replicate many acoustophoretic units to enable contactless manipulation on different scales. The mobile and scalable acoustophoretic system is expected to overcome several technical challenges, including a robust mechanical design, efficient wireless communication, and a battery-operated phased array of transducers (PAT) board.

In this work, we present AcoustoBots, a novel swarm robotic platform that introduces modularity and mobility to acoustophoretic systems, offering dynamic content presentation and interaction zones, to further explore the multimodal interactions of acoustophoresis in real environments (see Figure 1). These AcoustoBots are equipped with custom mini-PAT boards with wireless communication capabilities. AcoustoBots not only move freely on a tabletop or floor but can also change the direction and orientation of sound emission due to a hinge actuation system. In addition to creating a novel platform, we also reflect on the design, potential interaction scenarios, and implications, emphasizing that the technology push should be met with an interaction scenario pull. Consequently, they can dynamically position and adapt content in response to user interactions or specific scenarios. The integration of projected visuals with AcoustoBots' multimodal (haptics-audio-visual) capabilities creates a space where users can interact physically in a real environment that blends the multimodal interaction experiences such as touch, audio, and visual aspects seamlessly. To that end, we examined the design of our acoustophoretic swarm platform by starting with the individual design of each AcoustoBot and examining various combinations of components, their placements, and synergies to optimize performance and flexibility. Following this, we examined the collective dynamics through multimodal interaction scenarios in which AcoustoBots collaborate and work independently towards shared goals such as a swarm of haptic, audio, and levitation interactions. AcoustoBots can be utilized in a wide variety of real-world scenarios that require mobility, adaptiveness, and natural multimodal interactions, such as industrial delivery automation, large-scale room interaction in mixed reality, and multimodal installations in theme parks and museums.

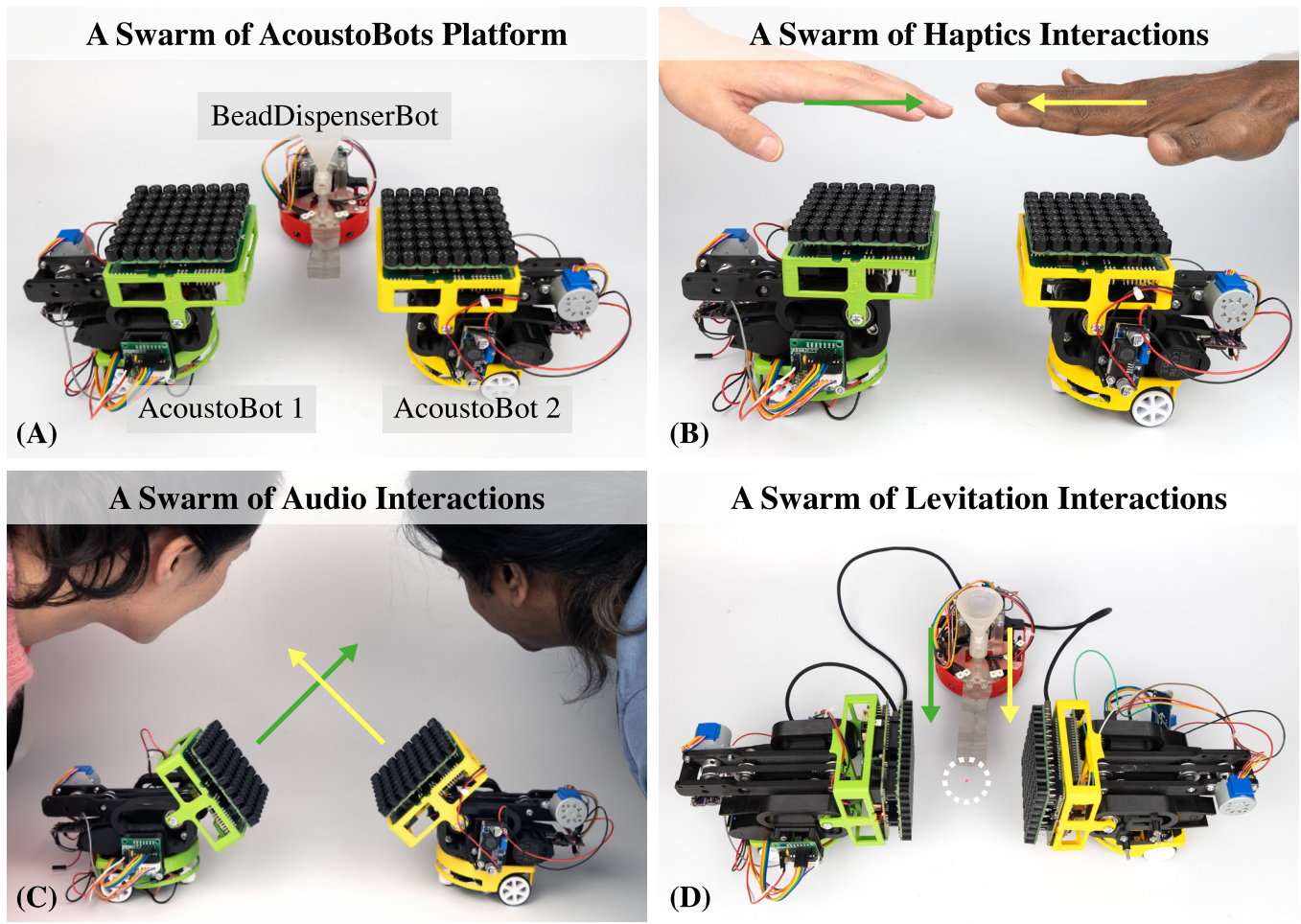

**Figure 1.** AcoustoBots: A unique and flexible swarm robotic platform that combines mobility with acoustophoretic functionalities, (A) AcoustoBots, comprised of 8 x 8 PAT boards, are mounted on mobile robots (left and right). Besides, a BeadDispenserBot (middle bot) automates the acoustic levitation process within our platform, (B) AcoustoBots can provide a swarm of haptics interactions independently, (C-D) AcoustoBots involve a dynamic hinge actuation system that can change the orientation of mounted PAT boards, allowing different spatial arrangements such as a swarm of audio, and levitation-based multimodal interactions, where users can interact physically in expansive environments.

The main contributions of this paper are:

- The design, development, and evaluation of AcoustoBots, a swarm robotic platform that combines mobility with acoustophoretic functionalities, offering multimodal interactions.
- An exploration of the design, both for individual AcoustoBot and their collective dynamics, detailing scenarios of a swarm of haptics, audio, and levitation.
- Our novel AcoustoBots design includes a hinge actuation system that controls the orientation of the mounted mini-PAT boards, enabling flexibility in the swarm of multimodal interactions.
- In addition, the design of a BeadDispenserBot, which can deliver particles to trapping locations, which automates acoustic levitation.

## 2 RELATED WORK

In this section, we review the progress in the acoustophoretic platform and a swarm of robots before we elaborate on how our AcoustoBots combine these fields to provide multimodal interactions.

### 2.1 Acoustophoretic Phased Array of Transducers

The ability to suspend particles at the nodes of a standing wave has been established for more than a century (Poynting and Thomson, 1904). The development of a PAT allowed for the arbitrary placement of local standing wave patterns. In turn, this led to the generated graphics using levitated particles (Omirou et al., 2015), (Sahoo et al., 2016), (ten Veen et al., 2018), although the voxels in these early works could only take certain predetermined positions. Later, holographic wavefront control techniques enabled the continuous placement and movement of levitated particles (Marzo et al., 2015), (Marzo and Drinkwater, 2019). Then, beyond levitation, the same principles have been extended to mid-air haptic feedback (Carter et al., 2013) and directional audio (Ochiai et al., 2017). These advances were later combined in multimodal acoustic trap displays (Hirayama et al., 2019), (Plasencia et al., 2020), where haptics, audio, and volumetric visual content, based on fast-moving levitated particles providing the PoV effect (Fushimi et al., 2019), were simultaneously generated by a single device.

This progress in terms of hardware and software, along with the versatility of acoustic levitation, has allowed the manipulation of all possible materials in interactive scenarios. Expanded polystyrene particles have been the most common rigid particles for acoustic levitation and have also been used as anchor points for levitating threads or lightweight fabrics, where visual content can be projected (Morales et al., 2019). In addition, levitating morsels of food (Vi et al., 2017) or droplets that encapsulate taste and smell (Vi et al., 2020) have also been proposed, demonstrating that acoustophoresis can provide experiences related to all senses. This wide range of delivered materials and modalities has rendered acoustophoresis useful for applications such as data physicalization (Gao et al., 2023). Closer to our work, Ultra-Tangibles was the first time that levitated materials were treated as tangible objects on a tabletop (Marshall et al., 2012), but interactions in this early approach suffered from the limitations of static transducer arrays.

### 2.2 Swarm of Robots for Multimodal Interactions

A swarm of robots involves multiple interactive elements that collectively behave to form unified interactions. These interactions are inspired by the swarm of robots, where the behavior of the group as a whole is influenced by individual members' actions. The swarm of interactions has many advantages: It offers collective interactions, visual appeal, adaptability to different contexts, and human-swarm interactions through gestures, touch, or other input methods. The swarm of interactions finds applications in various fields, including data visualization, artistic installations, and interactive media experiences. Several technologies offer swarm interaction approaches to leverage a swarm of robots for tangible interactions. For example, SwarmHaptics demonstrates the use of swarm robots for everyday haptic interactions such as notification, communication, and force feedback (Kim and Follmer, 2019). UbiSwarm displays information to users by collectively forming shapes or by their movements (Kim and Follmer, 2017). ShapeBots changes configuration to display information using shape-changing swarm robots (Suzuki et al., 2019). However, the large number of actuators needed to render a shape limits the resolution of these devices, making them complex, expensive, heavy, power-hungry, and limited in the coverage area. In addition, it limits real-time interaction and immersive experience.

The main limitations of the swarm robotic platforms mentioned above are the static transducer arrays and interaction volume that lack flexibility, which hinders the multimodal interaction experiences in a realistic environment. Recent work has extended the range of other types of haptic interfaces by integrating them onto robotic platforms such as HapticBots (Suzuki et al., 2021) is a novel encountered-type haptic approach for virtual reality based on multiple tabletop-size shape-changing robots, Hermits (Nakagaki et al., 2020) augment the robots with customizable mechanical add-ons to expand tangible interactions, HoloBots (Ihara et al., 2023) augment holographic telepresence with synchronized mobile robots, ShapeBots (Suzuki et al., 2019) enhance the range of interactions and expressions for tangible user interfaces, SwarmHaptics (Kim and Follmer, 2019) inform how users perceive and generate haptic patterns, and Zooids (Le Goc et al., 2016) interfaces comprised of autonomous robots that handle both display and interaction.

However, to the best of our knowledge, none of the previous work has investigated dynamically adjustable ultrasound phased arrays (i.e., changing both position and orientation) in a swarm platform or achieving mobile mid-air multimodal interactions in one system. Here, with wireless communication among modular transducer boards and a dynamic hinge actuation system, AcoustoBot, for the first time, enables large, flexible, scalable, mid-air multimodal interactions using a swarm of robots.

## 3 ACOUSTOBOT SYSTEM

### 3.1 AcoustoBot Definition and Scope

AcoustoBot is a self-actuated, movable, and flexible mini-PAT board that provides acoustophoretic multimodal interactions, such as haptic sensations, spatial audio, or visual artifacts via acoustic sound waves. Unlike all previous work on multimodal interactions (Marshall et al., 2012), (Hirayama et al., 2019), (Plasencia et al., 2020), AcoustoBot operates and communicates wirelessly. As a result, it can move around, delivering multimodal content. Users can interact directly with AcoustoBots by placing them on the tabletop or floor, and they can keep their hands right above them to enable hand tracking and follow functions. These features significantly extend the interaction area and thus dramatically increase the range of possible interactions, which is one of the main contributions of our work.

### 3.2 AcoustoBot Design

The AcoustoBot design comprises a self-actuated tangible robot, illustrated in an exploded view in Figure 2. The system's integral components encompass: 1) a battery-powered, flexible acoustophoretic phased array of transducers, which operates wirelessly, 2) a self-propelled Mona robot, which operates wirelessly and provides dynamic mobility for the mini-PAT board, 3) a dynamic hinge actuation system, which controls the orientation of the mini-PAT board using a geared hinge, and 4) a robust mechanical structural mount that unifies all these components.

We now delve into each of these components in detail.

#### 3.2.1 Acoustophoretic Phased Array of Transducers

In our AcoustoBot system, a customized, wirelessly operated, and battery-powered 8 x 8 mini-PAT board consists of a group of ultrasonic transducers, each operating at a single frequency of 40 kHz. These transducers are individually controlled by a field programmable gate array (FPGA) board, which is essential to generate the specific amplitude and phase patterns needed to sculpt the desired sound field. The key to achieving this control lies in the precise timing of when each transducer is activated or deactivated, as this determines the phase delay between them. For effective manipulation of sound, our system is updated at a minimum rate of 40 kHz, supporting up to 64 transducers.

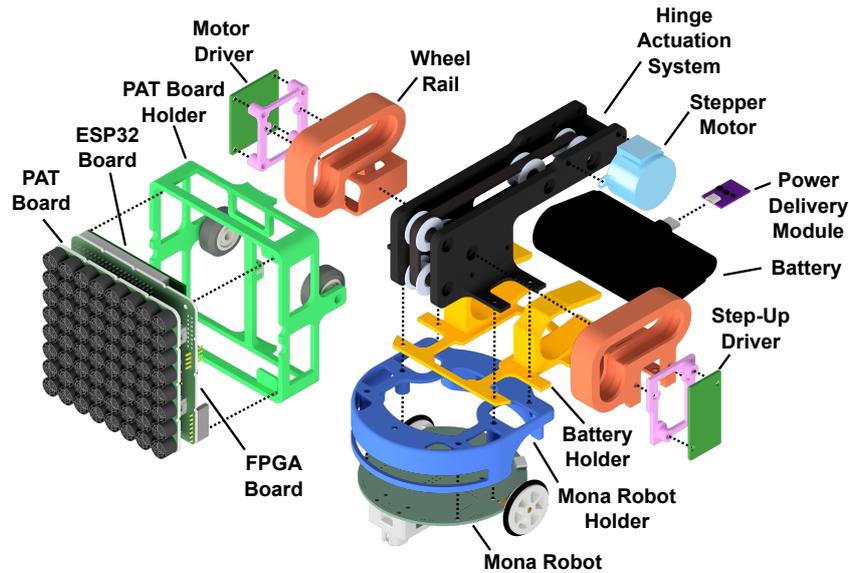

**Figure 2.** Exploded view of an AcoustoBot. Cable connections are not shown.

Compared to commercial systems, such as Ultraleap's mid-air haptics device, which can handle up to 256 transducers individually, this allows for a more complex sound field across a larger area (63 × 48 × 48 cm) (Faridan et al., 2022). However, the downside of using more transducers is the increased power consumption, which can affect the mobility of the system. One common solution to this issue is to use power banks, but this approach can make the system bulkier and less portable. Alternatively, there are smaller custom boards that use fewer transducers, such as TinyLev (Marzo et al., 2017), which simplifies operation by only allowing transducers to be turned on or off, resulting in a maximum of two-phase settings. This simplicity is achieved by connecting all transducers in one hemisphere, eliminating the need for individual controls. However, for more complex phase adjustments, a separate FPGA board would be necessary for each transducer. Although integrating an FPGA board adds to the design complexity, it is crucial to achieve better control over the sound field.

In contrast, our design features a compact and efficient setup centered around an 8 x 8 mini-PAT board. This setup includes a Waveshare CoreEP4CE6 on an Altera Cyclone IV FPGA board, an Adafruit ESP32 Feather V2 board, and a battery, as illustrated in Figure 3. This arrangement, with its modest number of transducers, is designed for low power consumption. It can be powered by a single 12V DC, 5000 mAh mobile battery, which balances power efficiency, control, and portability via a power delivery module (decoy). To ensure that the PAT board receives adequate power, we incorporated a step-up driver (power booster) that raises the battery output from 12 to 20V DC.

The operation of the system involves a server PC running a piston model, which generates motion parameters and transmits them as messages to the ESP32 using the user datagram protocol (UDP). These motion parameter messages include the calculated focal point, phase, and amplitude for each transducer element. These messages are then sent to the FPGA board (slave) from the ESP32 board (master) via the serial peripheral interface (SPI) protocol. Upon receiving these messages, the FPGA board generates square-wave drive signals up to 20V DC, which are then sent to the transducer array to create a controllable acoustic field. The phased array itself is designed as a one-sided square array (total weight is 300 gm:

phased array is 200 gm + battery weight is 100 gm), measuring 100 x 100 mm, with 8 x 8 transducers mounted on its surface. The structure of the array is 3D printed and houses 40 kHz piezoelectric ultrasonic transducers, each 10.5 mm in diameter, enabling precise control over the acoustic field generated.

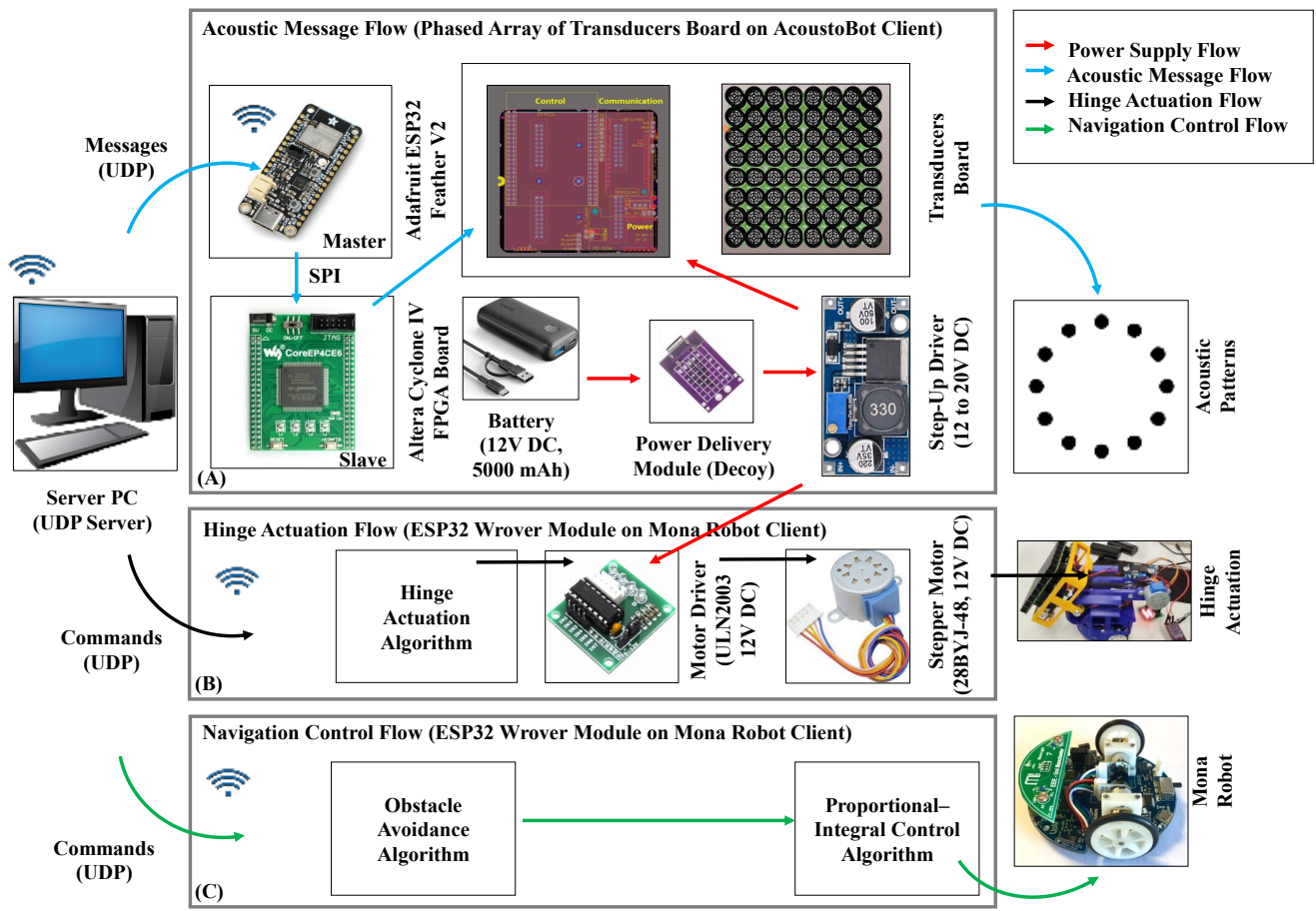

**Figure 3.** Data flow of an AcoustoBot. (A) Acoustic message flow, (B) Hinge actuation flow, and (C) Navigation control flow.

### 3.2.2 Self-Propelled Mona Robot

We chose the Mona robot[1] for this work because it is a compact, battery-operated, two-wheeled robot that can be controlled wirelessly. It also offers several advantages, including availability, speed, durability, and safety. The Mona robot is designed with an open-source, low-cost, flexible, and simple Arduino-based architecture that perfectly suits our research work. It includes an 80 mm diameter circular printed circuit board (PCB) mounted on two wheels, with a front axle supported by a resting wheel. It features five infrared (IR) sensors spaced at 35-degree intervals on the front half of the PCB. These sensors enable it to detect obstacles and other robots to avoid collisions.

### 3.2.3 Dynamic Hinge Actuation System

We connected the 28BYJ-48 12V DC stepper motor to the Mona robot (onboard ESP32 Wrover module) via the ULN2003 motor driver. The hinge activation of the Mona robot is controlled by the swarm robotic server PC over wirelessly. The server PC sends the control commands (activation/deactivation) to the hinge control client algorithm (on Mona robot) through UDP communication. The hinge control client algorithm turns the stepper motor in 1024 steps from 0 degrees (horizontal position) to a slope of 45 degrees (slope position), and 2048 steps to move the mini-PAT board from a slope of 45 degrees to 90 degrees (vertical position). Similarly, the hinge control client algorithm turns the stepper motor in a negative number of steps, such as -1024 steps, to move the mini-PAT board back from 90 degrees to 45 degrees and from 45 degrees to 0 degrees in -2048 steps, and reverses the motor's spin direction.

---

[1] https://github.com/ICE9-Robotics/MONA_ESP_lib/

### 3.2.4 Robust Mechanical Structural Mount

AcoustoBot is a custom robot made from modular components (as shown in Figure 2). At the base is a Mona robot for mobility, encased in a 3D-printed housing to protect components and allow access to IR sensors. Above the Mona robot is a 3D-printed case holding a Charmast ultra-compact battery (5000 mAh, 20 W, USB-C, 77.2 x 35.0 x 24.7 mm, 100 gm), powering the acoustophoretic mini-PAT board. Another 3D-printed part secures the mini-PAT board on top. We explored the two options for mounting the mini-PAT board on the Mona robot, 1) *Static configuration*: The square part holding the mini-PAT board can be detached and rotated in four positions, allowing the board to point in four directions. This setup enables two AcoustoBots to face their mini-PAT boards towards each other over a set distance for haptics and acoustic levitation, ensuring precise board positioning and orientation, and 2) *Dynamic configuration*: A subsequent design incorporates a dynamic board stage with a geared hinge actuation system, allowing electronic orientation adjustment. A single dynamic actuation system rotates the stage, enabling the mini-PAT board to switch between horizontal (0 degrees) for the swarm of haptics interactions, slope (45 degrees) for the swarm of audio interactions, and vertical (90 degrees) for the swarm of acoustic levitation interactions.

## 3.3 AcoustoBot Control, Navigation, and Communication System

### 3.3.1 Control System

We explore the design framework of AcoustoBot's control system, focusing on position and orientation. By applying the differential drive robot kinematic model (Klancar et al., 2017), we establish a control paradigm that ensures precise movement and efficient execution of its acoustophoretic multimodal interactions.

### 3.3.2 Navigation System

The AcoustoBot system employs a client-server architecture in which the core swarm control software operates as a UDP server, sending commands and messages to various UDP clients, including the AcoustoBots and BeadDispenserBots clients. To track the AcoustoBots' locations on the table or floor, we use a PhaseSpace[2] tracking system configured as a UDP client, enabling continuous monitoring of each bot's position and orientation.

### 3.3.3 Communication System

To establish effective communication within our AcoustoBot system, we developed a WiFi-based solution that integrates all components—the PC server (swarm control), AcoustoBot (Mona robot + mini-PAT board + hinge actuation) clients, and BeadDispenserBot (Mona robot + Bead dispenser) clients—into a unified client-server network. In this configuration, the PC functions as the central server, while AcoustoBots, PhaseSpace tracking system, and BeadDispenserBots operate as distributed clients. A router assigns unique IP addresses to each client based on their MAC addresses, ensuring clear identification and reliable communication across the network.

---

[2] https://www.phasespace.com/software/

## 3.4 BeadDispenserBot Design

We developed the BeadDispenserBot to supply particles to the AcoustoBots, as shown in Figure 4. The system includes a bead dispenser control algorithm on an ESP32 Wrover module on the Mona robot, a ULN2003 motor driver, a 28BYJ-48 5V DC stepper motor, and a 9V DC, 600 mAh Li-ion battery (see Figure 5). A server PC runs the control algorithm, sending activation and deactivation commands to the ESP32 Wrover module on the Mona robot. When activated, the ESP32 drives the stepper motor through four equal turns (out of 2048 steps), releasing four particles. Deactivation stops the motor, stopping the dispensing of particles.

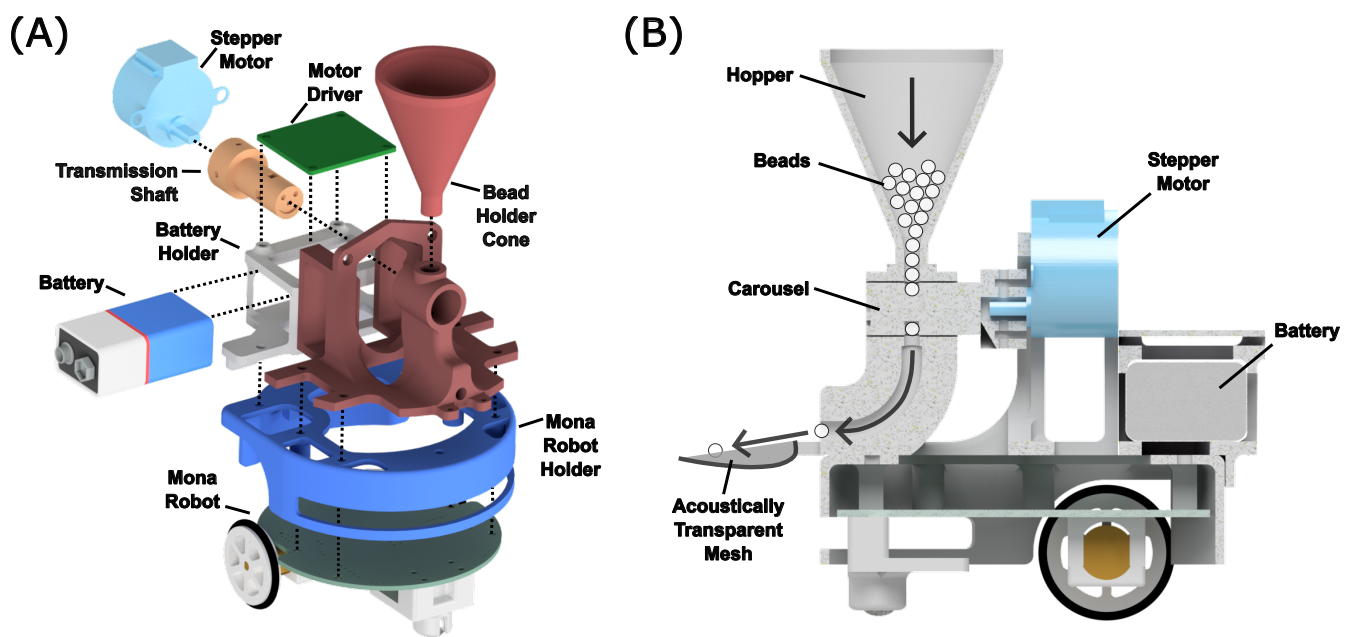

**Figure 4.** (A) Exploded view of a BeadDispenserBot. Cable connections and particle flow are not shown. (B) Cross-sectional diagram of the same bot showing how beads travel from the hopper, through the revolving carousel, which permits only one bead per quarter rotation of the motor shaft to be dispensed down the chute and into the acoustically transparent mesh bowl.

The server software for AcoustoBots and BeadDispenserBot, including acoustic control, hinge actuation, bots tracking, and bead dispense algorithms, was developed in C++ using Microsoft Visual Studio 2022 on Windows 11. The client software, implementing acoustic control, hinge actuation, bead dispense, obstacle detection, collision avoidance, and proportional-integral control, was written in C/INO using Arduino IDE 2.3.2 on Windows 11 and deployed on ESP32 Wrover and Adafruit ESP32 Feather V2 boards. FPGA code for AcoustoBots was developed in VHDL using Intel Quartus Prime Lite Edition 18.1 on Windows 11 and deployed on the Altera Cyclone IV FPGA board. For more information on the modules developed and the results captured, please refer to the AcoustoBots GitHub page[3].

---

[3] https://github.com/narsimlukemsaram/AcoustoBots/

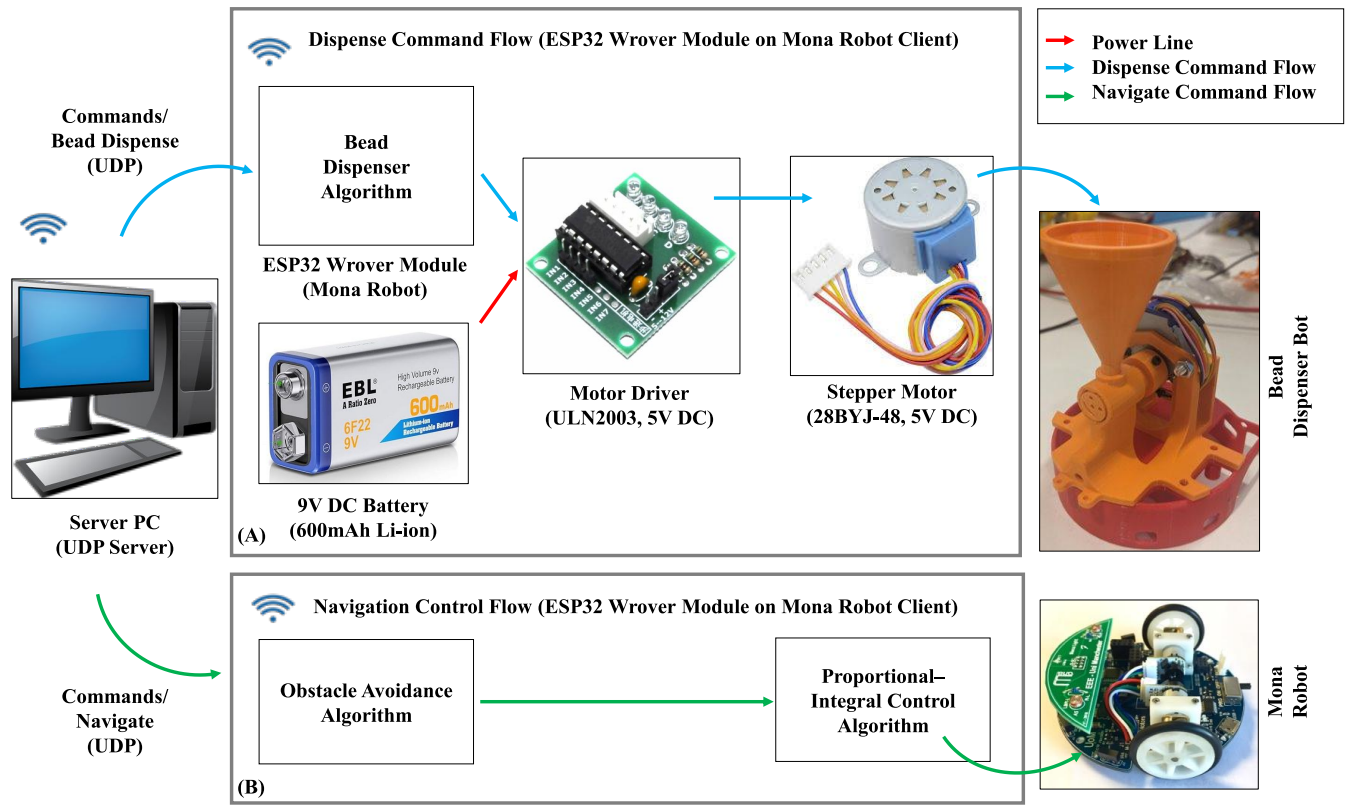

**Figure 5.** Data flow of a BeadDispenserBot. (A) Bead dispense flow, and (B) Navigation control flow.

## 4 SWARM OF ACOUSTOBOTS FOR MULTIMODAL INTERACTIONS

We faced several intricate challenges to realize our AcoustoBots system, in which mini-PAT boards operate, move, and communicate wirelessly. First, we developed battery-operated mini-PAT boards to avoid running off the main power supply, which is the case for conventional PAT boards. This was the first necessary step for making our platform cable-free and mobile. Second, we designed customized PAT boards to be small and lightweight so that they can be held by a 3D-printed housing that hinges when mounted on mobile robotic platforms. Following these, we fabricated housing units for all individual elements (battery, mini-PAT board, hinge, and robot) and assembled a compact AcoustoBot design, achieving a unique set of hardware attributes that balance such as power management and mechanical robustness. Finally, all of the above was architected into a common wireless-controlled system to support multimodal interactions with a swarm of robots.

### 4.1 Haptic Interactions with Swarm of AcoustoBots

The generation of mid-air haptics through ultrasound (Carter et al., 2013), (Obrist et al., 2013) has significantly improved shape rendering and tactile experiences in general (Obrist et al., 2015), (Shen et al., 2023, 2024), but current devices based on static arrays of ultrasonic transducers can only provide such experiences in a limited and fixed interaction area (Plasencia et al., 2020). In Figure 6, we illustrate AcoustoBots' new approach to large-area haptics, which leverages a swarm robotic platform and collaborates to overcome this challenge. Initially, AcoustoBots consisting of mini-PAT boards mounted on robots are randomly placed on the tabletop or floor, where the user's hands are also present (*approaching*) (see Figure 6A). A PhaseSpace tracking system provides information about the locations of AcoustoBots and the user's hands on the two-dimensional surface of the tabletop or floor. As a result, each robot can independently move and place a mini-PAT board beneath each of the user's hands (*aligning*) (see Figure 6B). Following this, mini-PAT board can deliver haptic content, for example, by using spatio-temporal modulation (by rapidly and periodically moving one or more focal points along the shape to be rendered). As the user moves their hands in different directions (note the arrows in Figure 6), AcoustoBots follow them to maintain the delivery of haptic content (*following*) (see Figure 6C).

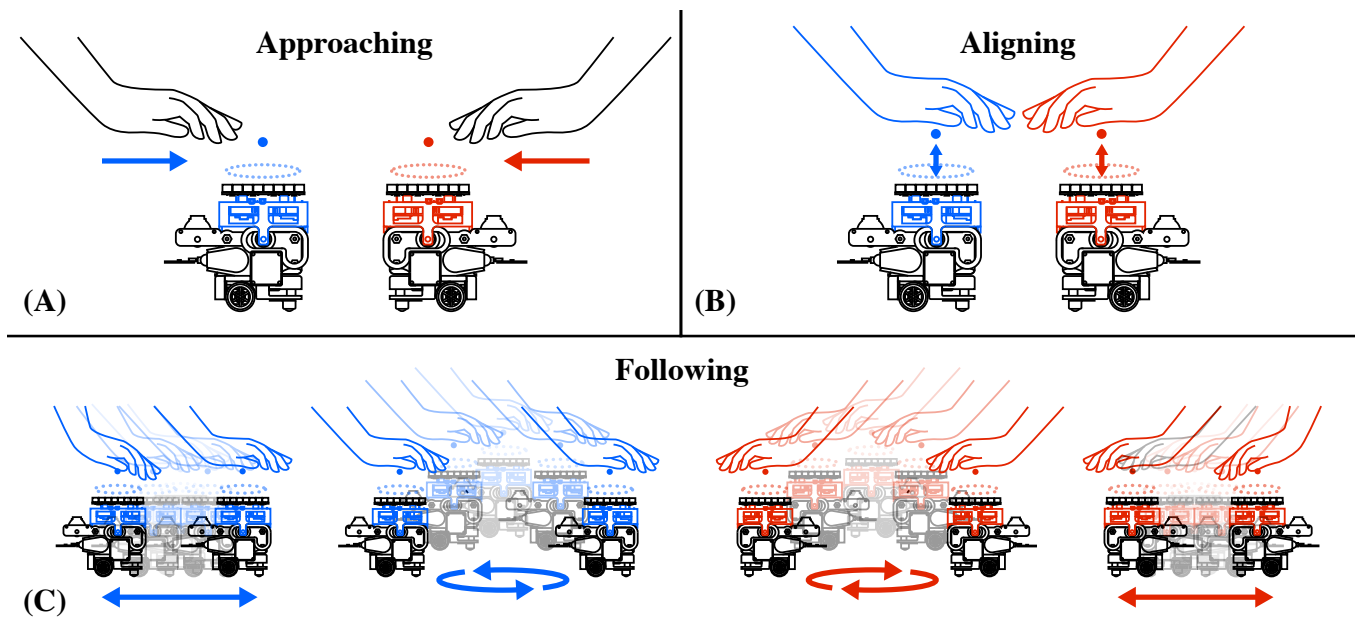

**Figure 6.** Independent haptic interactions with a swarm of AcoustoBots: (A) AcoustoBots and the user's hands have different initial positions (*approaching*), (B) Each AcoustoBot moves beneath a user's hand to deliver haptic content (*aligning*), and (C) As the user moves each of their hands in a different direction (see arrows), AcoustoBots can follow them and still provide haptic sensations (*following*).

## 4.2 Audio Interactions with Swarm of AcoustoBots

Similarly to haptic feedback, acoustophoresis can generate audible sound points from focused ultrasound waves, facilitating directional loudspeakers in the interaction space (Ochiai et al., 2017). This is achieved by modulating the audible signal in the ultrasonic carrier (Pompei, 2002). Initially, this was restricted to a single steerable audio column (Olszewski et al., 2005) and then expanded to multiple independent units emanating from the same transducer array system (Shi et al., 2015). In Figure 7, we illustrate how AcoustoBots, initially arranged for mid-air haptics as shown in the previous Section 4.1 and Figure 6, can rearrange themselves and provide audio for each user. Initially, AcoustoBots are randomly placed on the tabletop or floor, where the user's ears are also present (*approaching*) (see Figure 7A). A PhaseSpace tracking system provides information about the locations of AcoustoBots and the user's ears. As a result, each robot can independently move the mini-PAT board toward each of the user's ears (*aligning*) (see Figure 7B). Following this, mini-PAT boards can deliver audio content, for example, by using spatio-temporal modulation. As the user moves in different directions (note the arrows in Figure 7)), AcoustoBots follow them to maintain the delivery of audio content (*following*) (see Figure 7C).

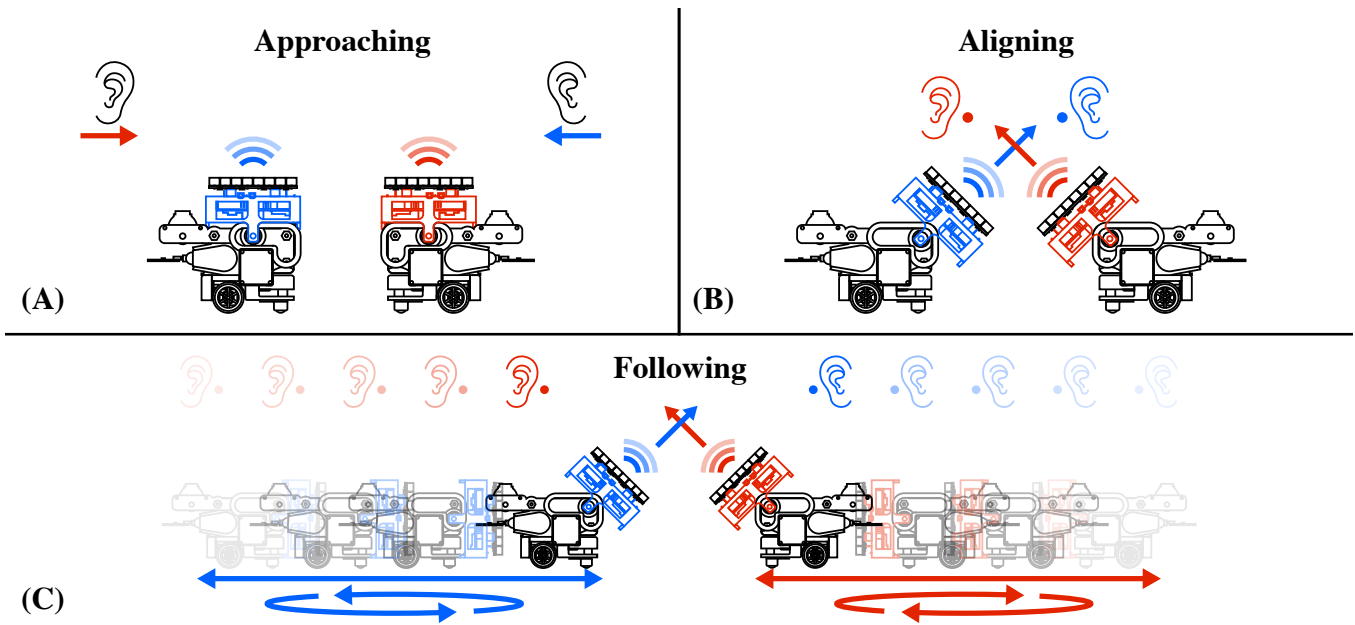

**Figure 7.** Collaborative audio interactions with a swarm of AcoustoBots: (A) AcoustoBots and the user's ears have different initial positions (*approaching*), (B) Each AcoustoBot inclines the mini-PAT board towards a user's ear to deliver audio (*aligning*), and (C) Each user moves in a different direction (see arrows), AcoustoBots can follow them and still provide audio feedback (*following*).

## 4.3 Visual Interactions with Swarm of AcoustoBots

For acoustic levitation (Marzo et al., 2015), (Marzo and Drinkwater, 2019), (Hirayama et al., 2019), (Plasencia et al., 2020), (Gao et al., 2023), AcoustoBots actively cooperate to position themselves at a predetermined distance and provide their own half of standing wave patterns where particles can be suspended. Limitations regarding the interaction area are not exclusive to haptics and audio, but rather an important issue in all interactive acoustophoretic applications. A significant feature of AcoustoBots is that they can switch between different arrangements and applications. In Figure 8, we illustrate how AcoustoBots, initially arranged for mid-air haptics and audio, as shown in the previous Sections 4.1 and 4.2 and Figures 6 and 7, can rearrange themselves and cooperate to generate standing wave patterns for acoustic levitation. For fully automated cooperative levitation, we consider an additional robot that is responsible for dispensing particles within the acoustic field generated by cooperating AcoustoBots. We refer to this robot as BeadDispenserBot. Initially, AcoustoBots are randomly placed on the tabletop or floor, where the BeadDispenserBot is also present (*approaching*) (see Figure 8A). A PhaseSpace tracking system provides information about the locations of AcoustoBots and the BeadDispenserBot. As a result, each robot can independently move and stay at an optimal distance (*aligning*) (see Figure 8B). Following this, AcoustoBots generate acoustic traps, while the BeadDispenserBot accordingly moves and dispenses a particle at the trapping location (*visualizing*) (see Figure 8C).

## 5 TECHNICAL EVALUATIONS

### 5.1 Experimental Setup

We conducted experiments on a custom-built AcoustoBots research platform (Figure 9), consisting of a Server PC (UDP server), a swarm router, two AcoustoBots (UDP clients), a PhaseSpace tracking system (UDP client), and a BeadDispenserBot (UDP client). The centrally controlled acoustic server software with the GS-PAT algorithm runs on a Server PC, which has a Windows 11, 64-bit OS, AMD Ryzen 7 5800H with Nvidia GeForce RTX 3060 GPU, and 16 GB RAM configuration. We used two identical AcoustoBots with acoustic control client software (along with hinge control, obstacle detection, collision avoidance, and robot control algorithms) and one BeadDispenserBot with bead delivery control client software (along with obstacle detection, collision avoidance, and robot control algorithms). We used PhaseSpace tracking system for motion tracking and navigating AcoustoBots and BeadDispenserBot clients. LED micro-drivers are attached to the AcoustoBots and BeadDispenserBot, to achieve accurate tracking and navigation in the test arena.

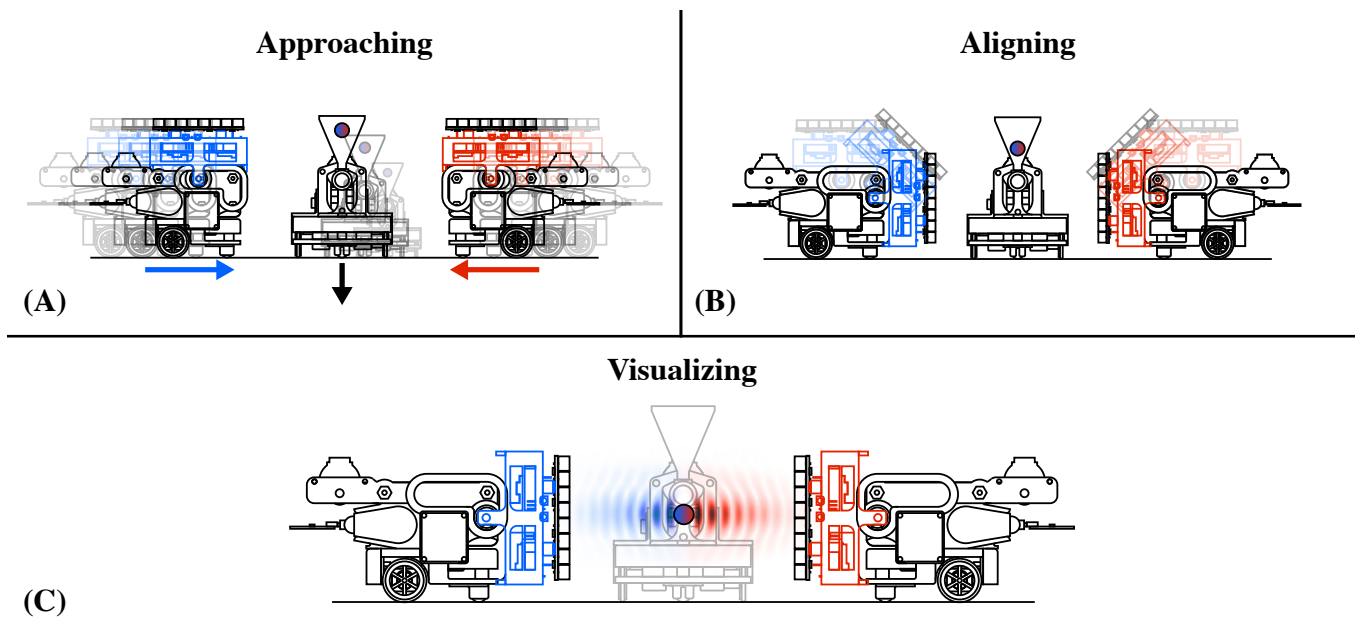

**Figure 8.** Cooperative visual interactions with a swarm of AcoustoBots: (A) In each AcoustoBot, the mini-PAT board is initially parallel to the carrying bot. An additional BeadDispenserBot is approaching for acoustic levitation (*approaching*), (B) Each AcoustoBot includes a hinge actuation system that rotates its mini-PAT board and places it vertically to the carrying bot (*aligning*), and (C) AcoustoBots generate acoustic joint levitation traps, while the BeadDispenserBot accordingly moves and dispenses a particle at the trapping locations (*visualizing*).

## 5.2 Experimental Results

We perform technical evaluations of the AcoustoBots and BeadDispenserBot to measure the accuracy and precision of hinge actuation and bead delivery in terms of position (in cm) and orientation (in degrees). In addition, we evaluated the focal point measurements in terms of pressure in Pascals (Pa) in the simulation and laboratory.

### 5.2.1 Accuracy and Precision of Hinge Actuation System

In this work, accuracy and precision are considered when evaluating the hinge actuation system. We evaluated the hinge actuation in terms of position and orientation errors by using the 28BYJ-48 12V DC stepper motor in full-step mode at 5 rpm. The hinge actuation was evaluated for each angle by driving it from 0 to 45 degrees (from horizontal to slope position), then commanding it from 45 to 90 degrees (from slope to vertical position) and recording the position and orientation by moving the hinge using a PhaseSpace tracking system (using rigidly attached LED markers on the PAT board) (see Figure 10A, 10B, and 10C). Position and orientation were measured by moving the hinge back to its original position. This process was repeated for a total of 10 times for hinge actuation, then another 10 times for deactuation, respectively. Figures 10D and 10E show that the accuracy (measured as mean absolute position and orientation error) is similar for horizontal and slope hinge actuation positions and slightly different for vertical positions, with a mean position error of ±2.50 cm and an orientation error of ±1.00 degrees. Based on Table 1 last column, we observe more precise (repeatable) behavior from the error bar on the horizontal and slope hinge actuation positions. This is probably due to a more even load distribution than the vertical hinge actuation position. Changing the array holding structural mounting to distribute loads around the vertical may improve precision (repeatability). Given the strong focal point (the optimal distance of a focal point is around 5 cm), these positioning and orientation errors are not expected to significantly impact the quality of the delivered haptics, audio, and levitation. However, if significant improvements in accuracy and precision were required, closed-loop position and orientation control would probably need to be implemented with the help of the camera, which would likely need to be implemented, impacting the accuracy, precision, and complexity.

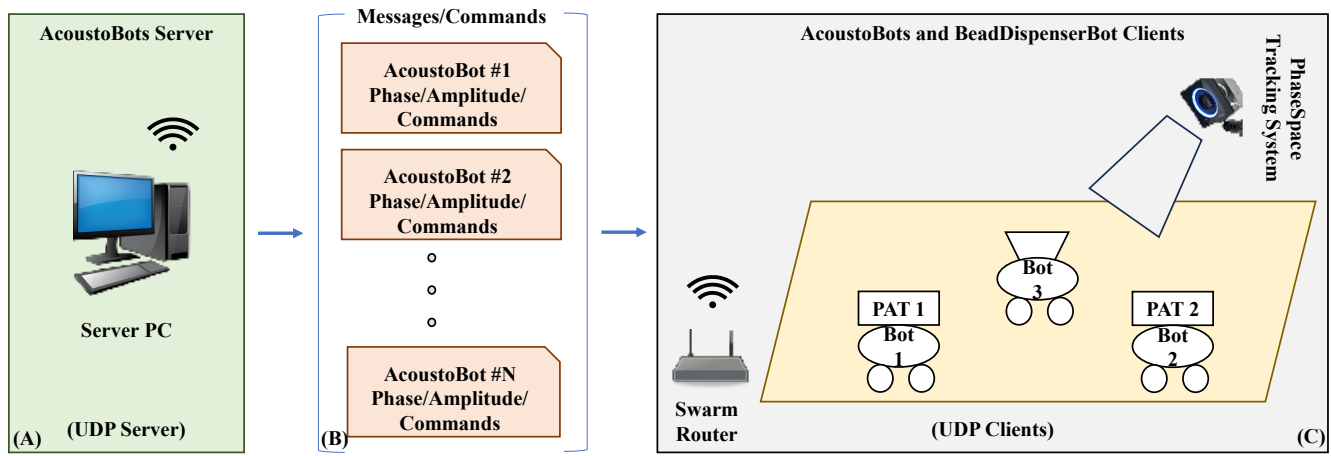

**Figure 9.** Experimental setup for a swarm of AcoustoBots evaluation, (A) It includes a Server PC (UDP Server), (B) Messages/Commands, (C) A swarm router, two AcoustoBots (UDP clients), a PhaseSpace tracking system (UDP client), and one BeadDispenserBot (UDP client).

Table 1 shows the results of the hinge responses obtained. The hinge actuation system behavior was found to be successful, with a mean position error of ±2.50 cm and an orientation error of ±1.00 degrees.

### 5.2.2 Accuracy and Precision of BeadDispenserBot

We evaluated the accuracy and precision of the BeadDispenserBot in terms of position and orientation errors by placing a bead precisely between the two AcoustoBots and collaboratively levitating the bead (see Figure 11A). We measured the position and orientation of the BeadDispnserBot to achieve the bead dispense automation process. This process was repeated a total of 10 times for the dispensing of the beads. A lightweight expanded polystyrene (EPS) bead (1 mm radius) was selected as the test particle due to its suitability for acoustic levitation. We plot a graph showing the reference and measured results of BeadDispenserBot in terms of both position (see Figure 11B) and orientation (see Figure 11C). Figures 11B and 11C show that the accuracy is good with a mean position error of ±0.45 cm and orientation error ±1.00 degrees. Based on Table 2 last column, we observe slightly more precise (repeatable) behavior of the BeadDispenserBot by levitating the particle. Changing the array holding the structural mounting to distribute loads around the vertical may improve precision (repeatability). Given the strong focal point (the optimal distance of a focal point is around 5 cm), these positioning and orientation errors are not expected to significantly impact the quality of the delivered levitation. However, if significant improvements in accuracy and precision were required, closed-loop position and orientation control would probably need to be implemented with the help of the camera, which would likely impact accuracy, precision, and complexity.

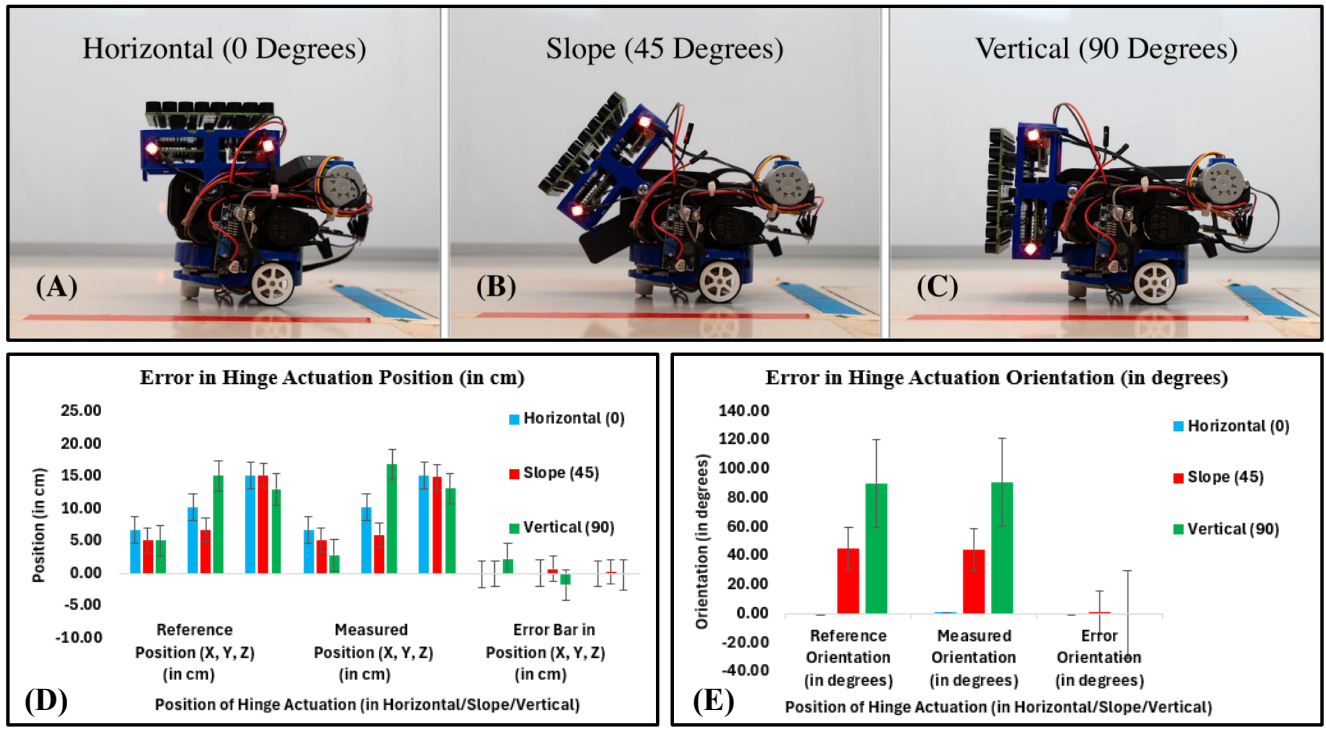

**Figure 10.** Accuracy and precision of hinge actuation system: (A) Horizontal position (0 degrees), (B) Slope position (45 degrees), (C) Vertical position (90 degrees), (D) Error in hinge actuation in terms of position (in cm). In the position error bar, the black line shows the mean (device accuracy), while the horizontal extent of the boxes shows precision (repeatability), and (E) Error in hinge actuation in terms of orientation (in degrees).

**Table 1.** Accuracy of the hinge actuation in terms of position and orientation errors.

| Position of Hinge Actuation (Horizontal/ Slope/ Vertical) | Reference Position (X, Y, Z) (in cm) and Orientation ($\psi$ in degrees) | | | | Measured Position (X, Y, Z) (in cm) and Orientation ($\psi$ in degrees) | | | | Error Position (X, Y, Z) (in cm) and Orientation ($\psi$ in degrees) | | | |
|---|---|---|---|---|---|---|---|---|---|---|---|---|
| | X | Y | Z | $\psi$ | X | Y | Z | $\psi$ | X | Y | Z | $\psi$ |
| Horizontal (0 degrees) | 6.70 | 10.20 | 15.10 | 0.00 | 6.78 | 10.21 | 15.13 | 0.38 | -0.08 | -0.01 | -0.03 | -0.38 |
| Slope (45 degrees) | 5.10 | 6.70 | 15.10 | 45.00 | 5.13 | 5.97 | 14.86 | 44.32 | -0.03 | 0.73 | 0.24 | 0.68 |
| Vertical (90 degrees) | 5.10 | 15.10 | 13.00 | 90.00 | 2.83 | 16.83 | 13.15 | 90.55 | 2.27 | -1.73 | -0.15 | -0.55 |

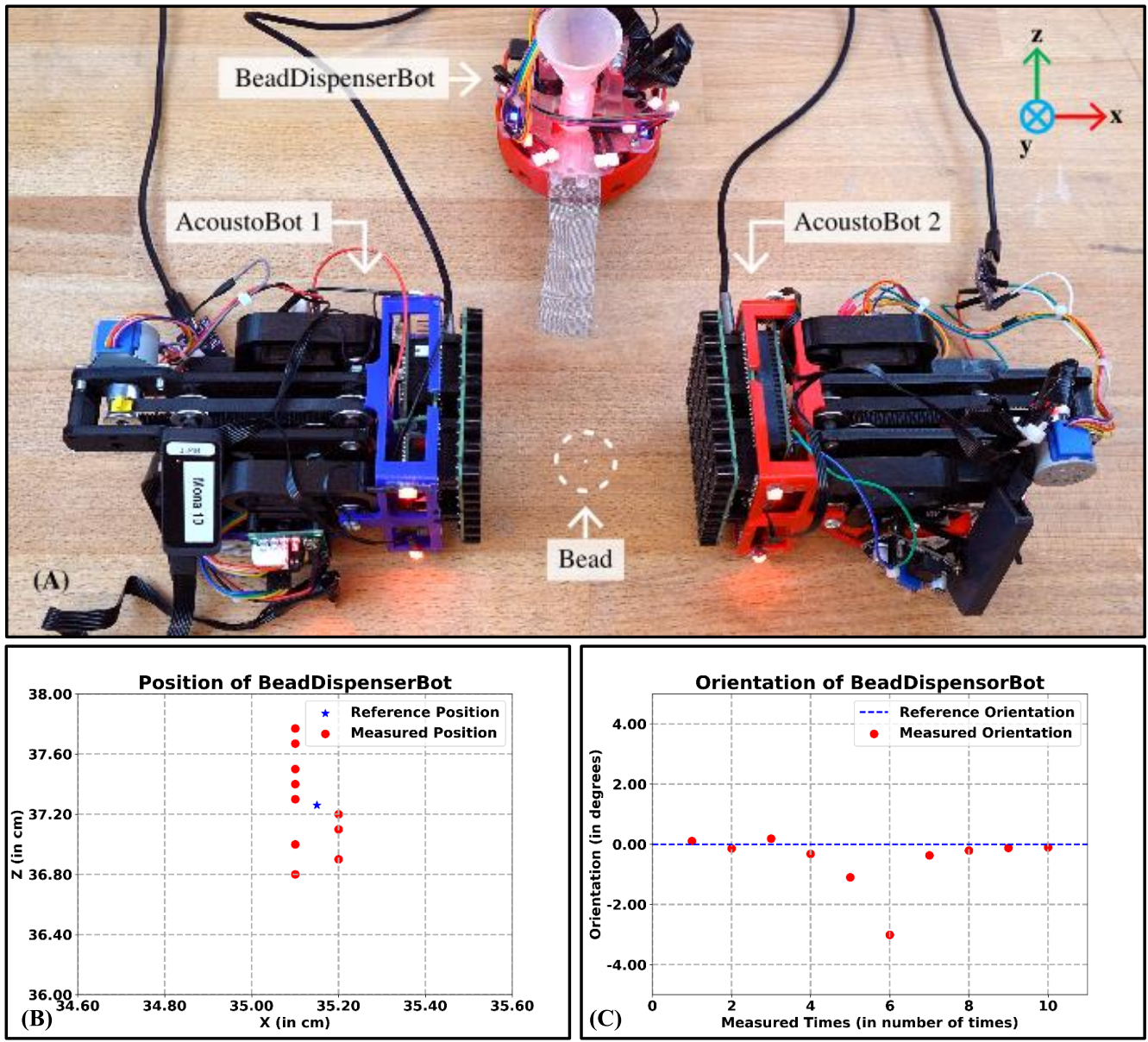

**Figure 11.** The accuracy of the BeadDispenserBot in terms of both the reference and measured position and orientation. (A) In the levitation scenario under the PhaseSpace tracking system, (B) A plot between the reference and measured position results of BeadDispenserBot (in cm), and (C) A plot between the reference and measured orientation results of BeadDispenserBot (in degrees).

Table 2 shows the results of the BeadDispnserBot obtained. The bead dispenser was found to be successful by levitating the particle with a mean position error of ±0.45 cm and an orientation error of ±1.00 degrees.

**Table 2.** Accuracy of the BeadDispenserBot in terms of position and orientation errors.

| S. No. | Reference Position (X, Y, Z) (in cm) and Orientation ($\psi$ in degrees) | | | | Measured Position (X, Y, Z) (in cm) and Orientation ($\psi$ in degrees) | | | | Error Position (X, Y, Z) (in cm) and Orientation ($\psi$ in degrees) | | | | Is Dispensed Bead Levitated? |
|---|---|---|---|---|---|---|---|---|---|---|---|---|---|
| | X | Y | Z | $\psi$ | X | Y | Z | $\psi$ | X | Y | Z | $\psi$ | |
| 1 | 35.15 | 4.40 | 37.26 | 0.00 | 35.10 | 4.40 | 36.80 | 0.11 | 0.05 | 0.00 | 0.46 | -0.11 | No |
| 2 | 35.15 | 4.40 | 37.26 | 0.00 | 35.20 | 4.40 | 36.90 | -0.14 | -0.05 | 0.00 | 0.36 | 0.14 | Yes |
| 3 | 35.15 | 4.40 | 37.26 | 0.00 | 35.10 | 4.30 | 37.00 | 0.19 | 0.05 | 0.10 | 0.26 | -0.19 | Yes |
| 4 | 35.15 | 4.40 | 37.26 | 0.00 | 35.20 | 4.35 | 37.10 | -0.32 | -0.05 | 0.05 | 0.16 | 0.32 | Yes |
| 5 | 35.15 | 4.40 | 37.26 | 0.00 | 35.20 | 4.30 | 37.20 | -1.10 | -0.05 | 0.10 | 0.06 | 1.10 | No |
| 6 | 35.15 | 4.40 | 37.26 | 0.00 | 35.10 | 4.40 | 37.30 | -3.01 | 0.05 | 0.00 | -0.04 | 3.01 | No |
| 7 | 35.15 | 4.40 | 37.26 | 0.00 | 35.10 | 4.40 | 37.40 | -0.37 | 0.05 | 0.00 | -0.14 | 0.37 | Yes |
| 8 | 35.15 | 4.40 | 37.26 | 0.00 | 35.10 | 4.40 | 37.50 | -0.21 | 0.05 | 0.00 | -0.24 | 0.21 | Yes |
| 9 | 35.15 | 4.40 | 37.26 | 0.00 | 35.10 | 4.40 | 37.67 | -0.12 | 0.05 | 0.00 | -0.41 | 0.12 | Yes |
| 10 | 35.15 | 4.40 | 37.26 | 0.00 | 35.10 | 4.40 | 37.77 | -0.10 | 0.05 | 0.00 | -0.51 | 0.10 | No |

### 5.2.3 Focal Point Measurements

The multimodal performance of the swarm of Acoustobots is analyzed by MATLAB simulations and validated by scanning the generated pressure field for the three different scenarios using microphone measurements in the laboratory, such as i) *Scenario 1*: Independent haptic focal points with a swarm of AcoustoBots (see Figure 12), ii) *Scenario 2*: Collaborative audio focal points with a swarm of AcoustoBots (see Figure 13), and iii) *Scenario 3*: Cooperative visual focal point with a swarm of AcoustoBots (see Figure 14).

Table 3 shows the results of the simulation and microphone measurements obtained. Key insights emerge from these results:

In *Scenario 1*, both AcoustoBots operated independently at a focal point, achieving a simulated pressure of 4469.90 Pa each, while microphone measurements recorded 2956.80 Pa and 2941.00 Pa for AcoustoBots 1 and 2, respectively. This discrepancy likely stems from practical limitations like acoustic losses, calibration variations, or slight misalignment, though the consistency between the two AcoustoBots confirms simulation predictions and microphone measurements are similar.

In *Scenario 2*, both AcoustoBots focused at a shared focal point, yielding a simulated maximum of 2791.30 Pa, with a microphone maximum of 2665.40 Pa. This reduction is likely due to destructive interference between the sources, and the close simulation-experiment alignment suggests the model effectively accounts for such interactions.

In *Scenario 3*, joint focus produced higher pressures, with 7106.10 Pa simulated and 5675.50 Pa microphone measured results, and it is possibly due to optimized alignment and stronger constructive interference. Both the simulation results and lab measurements clearly indicate that the maximum acoustic pressure is located at the same point, confirming the design's effectiveness and accuracy. The larger simulation-experiment deviation here implies that higher pressures may amplify nonlinear or environmental damping effects in real measurements.

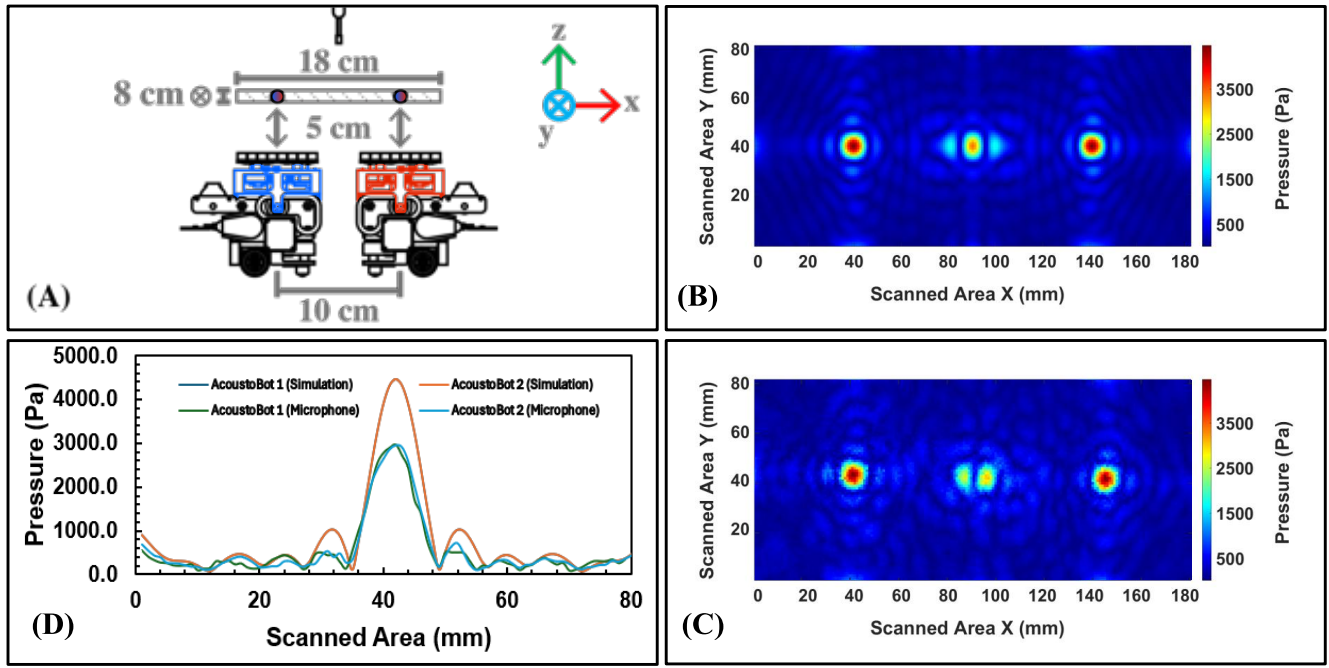

**Figure 12.** Focal point measurements for *Scenario 1*: (A) Horizontal setup (0 degrees) for independent haptics feedback, (B) Experimental results in simulation using MATLAB, (C) Experimental results in measurements lab using a microphone, and (D) A plot between the simulation and microphone results along vertical axis.

**Table 3.** Measurements in Simulation (MATLAB) vs Measurements in Lab (Microphone).

| Simulation in MATLAB/ Microphone Measurements in Lab | Scenario 1 | | Scenario 2 | Scenario 3 |
|---|---|---|---|---|
| | AcoustoBot 1 (Maximum Pressure at Focal Point) (in Pascals) | AcoustoBot 2 (Maximum Pressure at Focal Point) (in Pascals) | AcoustoBot 1 and 2 (Maximum Pressure at Shared Focal Point) (in Pascals) | AcoustoBot 1 and 2 (Maximum Pressure at Joint Focal Point) (in Pascals) |
| Simulation | 4469.90 | 4469.90 | 2791.30 | 7106.10 |
| Microphone | 2956.80 | 2941.00 | 2665.40 | 5675.50 |

For every scenario, the full width at half maximum (FWHM) of the focal point was computed in order to evaluate the beam confinement. The consistent finding of the FWHM to be almost equal to one wavelength ($\lambda$) indicates a tightly focused and well-confined acoustic beam. This shows how effectively the system generates precise sound-field control.

## 6 POTENTIAL ACOUSTOBOTS INTERACTION SCENARIOS

Here, we demonstrate the various potential interaction scenarios of AcoustoBots (see Figure 15).

In haptic interactions, amplitude modulation allows for precise control of stimulation at a specific point on the single hand (see Figure 15A), enhancing tactile feedback for haptic systems. Multipoint spatio-temporal techniques enable dynamic control across multiple hand regions (see Figure 15B), facilitating complex tactile feedback. Additionally, targeted tactile contact on the lips offers distinct sensations useful for haptic communication (see Figure 15C).

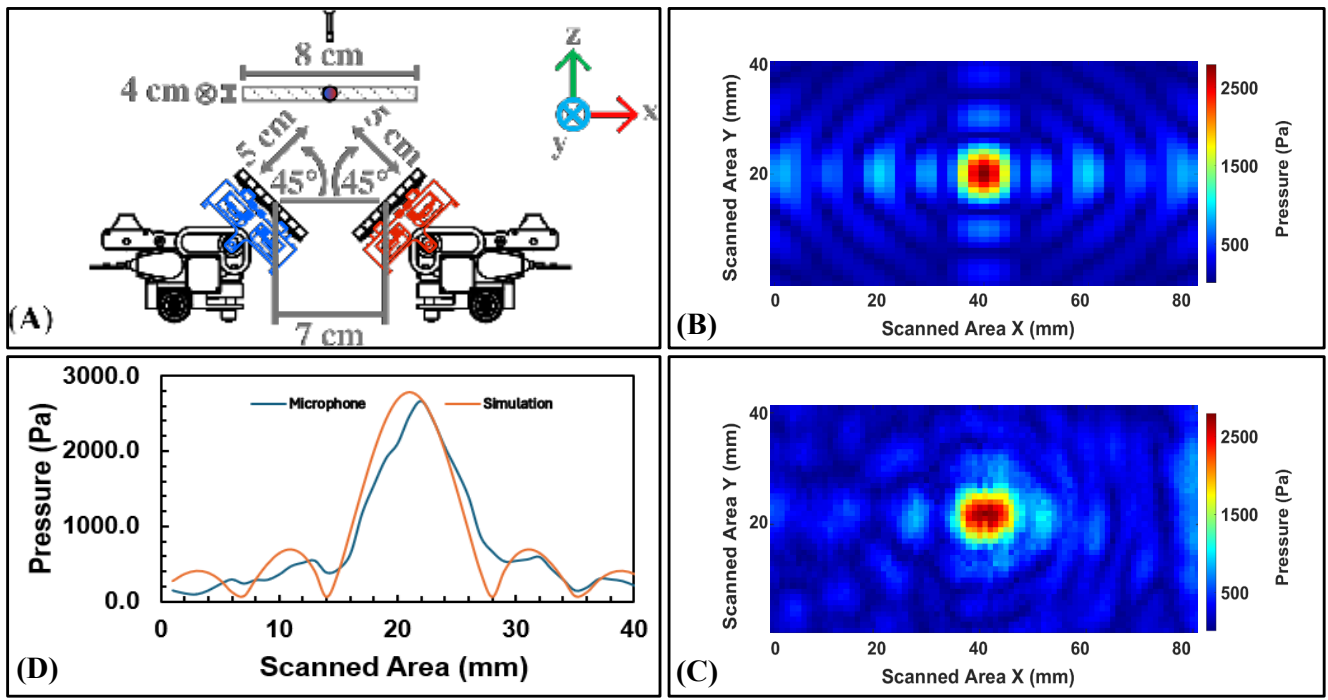

**Figure 13.** Focal point measurements for *Scenario 2*: (A) Slope setup (45 degrees) for collaborative audio feedback, (B) Experimental results in simulation using MATLAB, (C) Experimental results in measurements lab using a microphone, and (D) A plot between the simulation and microphone results along vertical axis.

In audio interaction, the audio spotlight focuses sound into a narrow beam, similar to how a spotlight focuses light (see Figure 15D), and it is helpful in museums and retail stores. Similarly, the directional audio creates a narrow, directed sound beam that can be clearly heard within its path but remains inaudible outside of it (see Figure 15E).

For levitation interaction, single-sided setups manipulate single particles (see Figure 15F), while double-sided fields control multiple particles for applications like sorting (see Figure 15G), and wall reflectors create standing waves for object levitation (see Figure 15H). Techniques for the handover of particles, either single- or double-sided, enable precision in non-invasive handling and microassembly (see Figure 15I).

# 7 DISCUSSIONS

In the field of swarm robotics, acoustophoresis introduces a novel way to visualize interactions and share information with robots, enhancing the efficiency of interactions (Le Goc et al., 2016), (Suzuki et al., 2019), (Kim and Follmer, 2019), (Suzuki et al., 2021), (Ihara et al., 2023). At the same time, multimodal interactions have sparked interest among the swarm robotics community because of their theoretical benefits and the novel interaction opportunities they provide (Marshall et al., 2012), (Hirayama et al., 2019), (Fushimi et al., 2019), (Plasencia et al., 2020), (Gao et al., 2023).

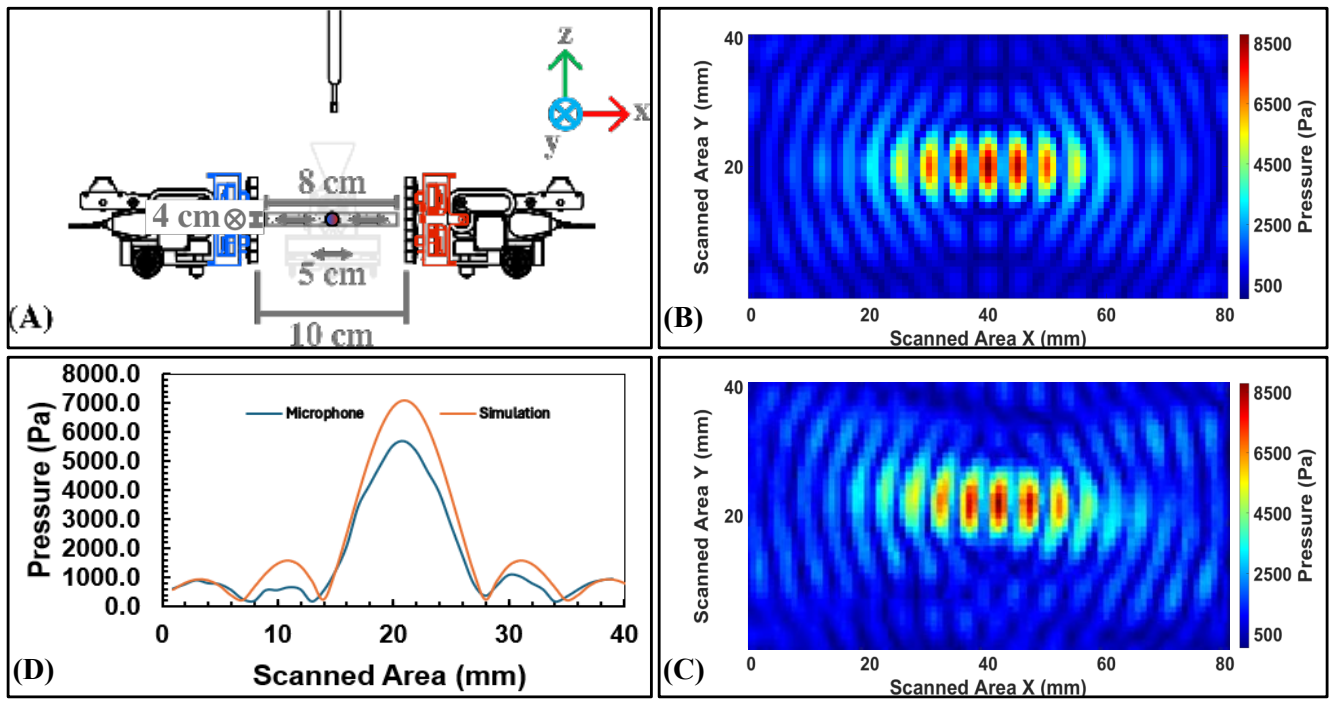

**Figure 14.** Focal point measurements for *Scenario 3*: (A) Vertical setup (90 degrees) for cooperative visual focal point with a swarm of AcoustoBots, (B) Experimental results in simulation using MATLAB, (c) Experimental results in measurements lab using a microphone, and (D) A plot between the simulation and microphone results along vertical axis.

To explore multimodal interactions using acoustophoretic phased arrays, we developed AcoustoBots, mobile miniature robots equipped with mini-PAT boards and wireless communication. Unlike static arrays, AcoustoBots move freely, adapting to user interactions or scenarios. By integrating multimodal interactions, this approach eliminates bulky head-mounted displays, offering an intuitive and seamless multimodal experience. Here, we explore the promising possibilities enabled by AcoustoBots while also outlining the accompanying limitations and challenges that emerge.

## 7.1 Modularity and Scability

In traditional static, singular, fixed PAT boards, mobility is limited by external power supply and fixed wired connections, making immersive experience challenging (Marshall et al., 2012), (Hirayama et al., 2019), (Fushimi et al., 2019), (Plasencia et al., 2020), (Gao et al., 2023). Here, we introduced AcoustoBots, a solution that tackles these challenges with battery-operated plug-and-play mini-PAT boards integrated with Mona robots. These AcoustoBots communicate wirelessly, enhancing both modularity and mobility. These AcoustoBots are also versatile for various applications and rich immersive experience with wireless synchronization, which enables multiple mini-PAT boards to collaboratively generate standing waves, focal traps, haptic feedback, and audio points. A hinge actuation system further enhances flexibility in response to user and environmental changes. Consequently, dynamically position and adapt content in response to user interactions or specific scenarios. Our work showed that the deployment of AcoustoBots enables large-scale complex interaction tasks, with swarm robotics significantly improved the multimodal interactions. These include independent haptic feedback for users hands, collaborative audio interactions for users ears, and cooperative visualization interactions between paired AcoustoBots.

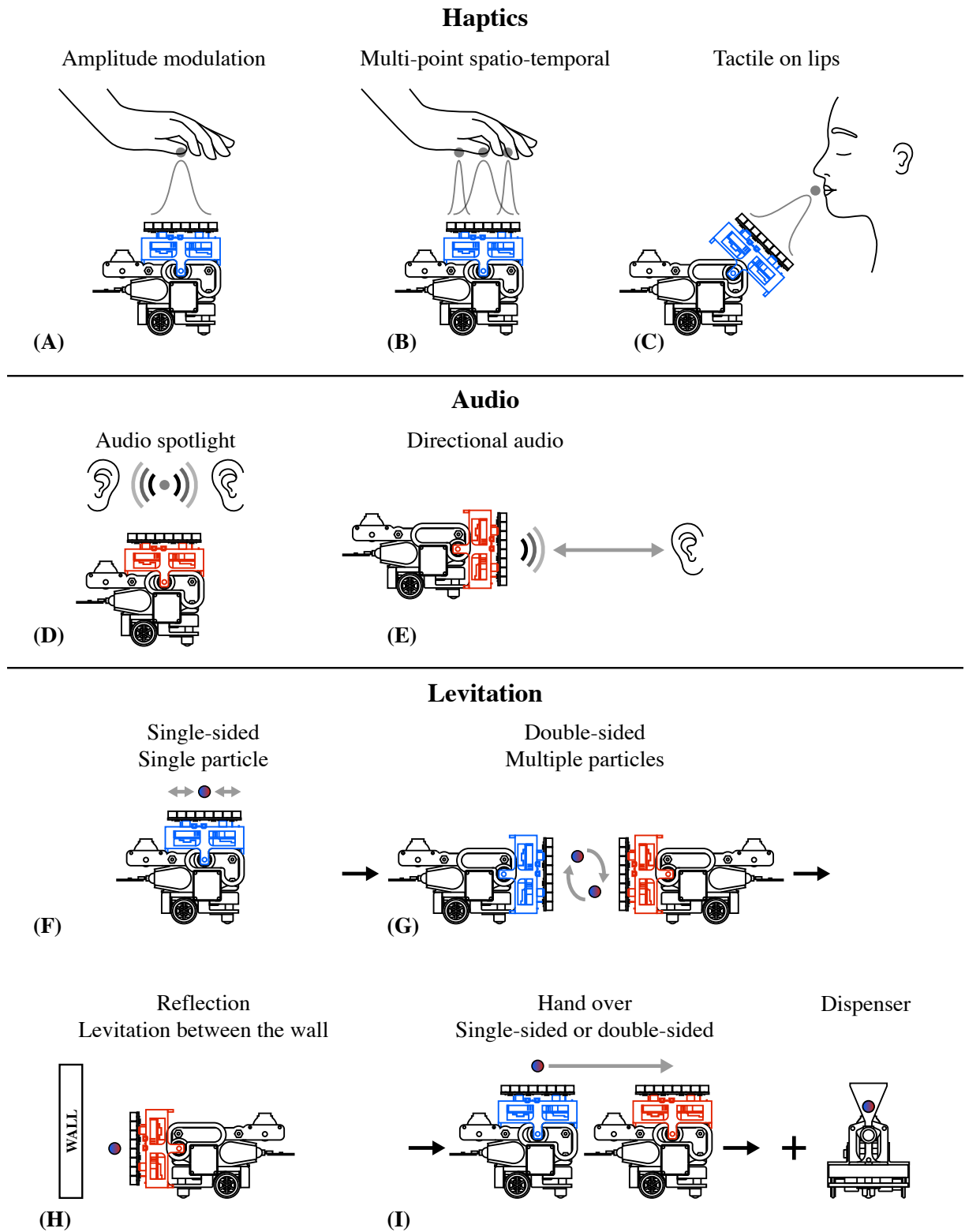

**Figure 15.** Potential AcoustoBots interaction scenarios. (A)-(C) for haptics interactions, (D)-(E) for audio interactions, and (F)-(I) for levitation interaction scenarios.

Although current experiments demonstrate coordination between two robots, the proposed system is designed with scalability in mind to support a large, fully distributed swarm of acoustic robots. Transitioning from two robots to a fully distributed swarm of acoustic robots introduces several technical challenges to be addressed in the future, including decentralized control (robot control system and acoustic field generation), communication constraints, and clock synchronization between multiple agents. Our current work is laying the groundwork for future deployment toward a fully distributed acoustic swarm system with a larger number of robots. This swarm system could work in unison to create dynamic and interactive environments, significantly enhancing the immersive user experience in mixed-reality applications. However, it may also cause interference in wireless communication, which can lead to increased computational demands. From a futuristic perspective, the tilting mechanism could be enhanced by integrating a telescopic lift, allowing each mini-PAT board to move vertically in addition to its tilting capabilities. This dual functionality would provide significantly greater flexibility and adaptability.

## 7.2 Multimodal Interactions

We envision that our swarm robotic platform can greatly benefit the multimodal interactions in room-scale mixed-reality applications. For instance, users are not constrained by fixed interaction space, like (Kim and Follmer, 2019), (Suzuki et al., 2021) and have interaction in rich scenarios like museums, galleries, and collaborative workspace. They can move and interact with any digital and physical content in the room. Meanwhile, the tracking system notify the AcoustoBots platform. A swarm of AcoustoBots follow and respond to user action and provide situational haptic feedback, directional audio feedback, and visual enhancement according to location and content for immersive experiences. Similarly, in our tabletop or floor setup, acoustophoretic phased arrays enable flexible mid-air multimodal interactions via Mona robots and hinge actuation, with horizontal-plane freedom for lateral movement and rotation. With enhanced vertical freedom, there be fewer constraints on multimodal interactions on the tabletop or floor, enabling more dynamic activities such as multi-user collaboration for education, planning complex projects, and simulating real-world scenarios with spatial flexibility.

Looking beyond the current technical implementation, more future work can explore new application opportunities enabled by AcoustoBots. Inspired by design futuring used in human-computer interaction (HCI) research (Beşevli et al., 2024), we organized an ideation workshop focusing on content creation, such as collaborative mapping on a tabletop or floor. Integrating multiple modalities (vision, hearing, touch) can enable the creation of novel multisensory interactions with content and enhance user immersion and engagement with content. Design futuring methodologies guide participants in experiencing crafted scenarios, fostering reflection (Kozubaev et al., 2020) on the diverse possibilities of multisensory experiences (Velasco and Obrist, 2020) where the human senses meet acoustophoresis, enabling novel self-actuating, multimodal robotic swarm interactions. This approach also allows reflection on design responsibilities and possible ethical considerations (Cornelio et al., 2023) across social, public, and private contexts. Looking ahead, we aim to explore broader use scenarios through future-thinking workshops involving multiple external stakeholders and carrying out feedback from user studies. This help to expand the design opportunities around AcoustoBots, and assessing potential benefits and challenges for multimodal interactions.

## 7.3 Limitations and Future Goals

Our initial experiments with AcoustoBots included limited multimodal interactions, such as a swarm of haptics, audio, and levitation interactions, using a setup of two robots, laying the groundwork for future deployment in a larger swarm of robots. Future work aims to incorporate additional modalities, such as taste and smell, for richer, multimodal experiences. We also plan to conduct formal user studies to evaluate AcoustoBots' effectiveness in specific data physicalization applications. Although this work established a distributed but centrally controlled algorithm for computing the phase and amplitude of transducers, developing a fully distributed approach supports a larger swarm of robots without losing communication efficiency. To enable wireless synchronization, we explore software-based synchronization over current hardware-based synchronization.

## 8 CONCLUSION

In this work, we presented AcoustoBots, a novel platform that combines acoustophoretic multimodal interactions with swarm robotic scenarios, significantly extends the interaction space of acoustophoretic applications. AcoustoBots are assembled by miniature phased array of ultrasonic transducer boards that are mounted on a swarm of robots, along with a hinge actuation system that adjusts the orientation of the phased array board. In contrast to conventional implementations where acoustophoretic units are static, AcoustoBots can move in space while delivering multimodal content, or reconfigure themselves to swap between applications. Designed into a common wireless controlled system to support swarm cooperative and collaboration functions, AcoustoBots offers the integration of currently restricted experiences in realistic multimodal interaction environments.

## ACKNOWLEDGMENTS

This work was supported by the EPSRC Prosperity partnership program - Swarm Spatial Sound Modulators (EP/V037846/1), by the Royal Academy of Engineering through their Chairs in Emerging Technology Program (CIET 17/18), EU H2020 research and innovation project Touchless (101017746), and by the UKRI Frontier Research Guarantee Grant (EP/X019519/1). The authors also thank Ben Kazemi for his contribution to designing the wireless mini PAT board, Ana Marques for her support in creating the images and videos, and James Hardwick for his voice-over for the accompanying video.